# IEEE Copyright Notice

# 3D Camouflaging Object using RGB-D Sensors

Ahmed M. Siddek[1], Mohsen A. Rashwan[2], Islam A. Eshrah[2]

[1]Suez University, Suez, Egypt
[2]Cairo University, Giza, Egypt
ahmed.siddek@suezuni.edu.eg, mrashwan@rdi-eg.com, isattar@eng.cu.edu.eg

**Abstract**— *This paper proposes a new optical camouflage system that uses RGB-D cameras, for acquiring point cloud of background scene, and tracking observers' eyes. This system enables a user to conceal an object located behind a display that surrounded by 3D objects. If we considered here the tracked point of observer's eyes is a light source, the system will work on estimating shadow shape of the display device that falls on the objects in background. The system uses the 3d observer's eyes and the locations of display corners to predict their shadow points which have nearest neighbors in the constructed point cloud of background scene.*

## I. Introduction

In this study, a new camouflage system is proposed for homeland security or military applications. This system can work on concealing military equipment in order not to be detected by enemies. To the author knowledge, it is the only one which based on tracking the observer's eyes using RGB-D sensors. These sensors are located in the side of the object required to be hidden, rather than the observer side. It needs no preprocessing environmental steps. The proposed system is used in indoor environment due to the limitations of the infrared depth sensor and the display device.

In this paper, we present a methodology of determining the camouflage images depends on estimating the shadow shape. If a light source that located at certain 3D point and focused on camouflage object covered by LCD, is substituted by a camera pointed at the same object and located at the same 3D point, the shadow of the display on the background (BG) will define the occluded region which is used to extract the camouflage image. And if the background scene is not planar as it has 3D objects, the shadow will deform and take the shape of objects that enclose it at each observer's location. We briefly review related work in section (II) and then provide system description and components that outlines the remainder of the paper.

## II. previous work

The current implemented camouflage systems are presented here. Optical camouflage system that uses RGB-D sensors and LCD [1], computes the quantized depth levels of background scene to camouflage 2D object of any shape and dimension that is located behind LCD. Its constraints are: (1) screen surface must be located parallel to the XY planes of the two kinect sensors, and (2) it must also be parallel to the background depth planes. The system [2, 6] uses Retro-reflective Projection Technology (RPT). Its limitations are the imperfect transparent of the covered object, the use of half-mirror in front of the observer's eyes, the existence of a projector in the observer side either on the ground or mounted on his head, and the projected image is not projected on the retroreflective material only but also on other surrounding objects. Some methods generate random or manual background patterns that surround the camouflaged object [3]. Other, use projector-camera system [4] with front and back cameras. The back camera is placed behind a planar panel that displays the projected image. It carries out image transformations using homogenous matrix (H) and projective transformation. Some drawbacks of this approach, are the misalignment on the boundaries and severe distortion for the 3D dynamic background. The paper [5], works on the problem by capturing the scene without the object from many viewpoints (approximately 10-25 images) and using background matching algorithms then the best surface pattern is printed on a cube-shaped object. Its advantage is hiding a 3D object from many views simultaneously. But its limitation is that it must first capture the scene without the object to collect more than 10 images of the surroundings.

## III. Proposed Camouflage System

### A. Algorithm Components

The first two components, are acquiring the point cloud of BG scene using $K_b$ sensor and obtaining real-time observation point using $K_f$ sensor. The observation point (OP) is derived in [1], except that the depths of eyes are calculated simply from depthmap of $K_f$. The procedures of the other two components in the algorithm are: (1) predicting the outer-boundary shadow of camouflage object that is covered by LCD at each 3D head point or 3D OP using OpenNN library, and (2) searching the nearest point in scene point cloud to the predicted shadow points using nearest neighbor algorithm of *kd-tree*.

To predicate the 3D shadow shape, a technique which depends on ray tracing that intersects the camouflaged object (CO) is used. The CO here, is the LCD display which has corners or vertices (control points). This technique uses observation point (OP) and the locations of display corners to predict their shadow points which have nearest neighbors in the constructed point cloud of BG. Fig. 1 shows a block diagram that illustrates the steps require to conceal the LCD by predicting the shadow points.

### B. Point Cloud

It represents the external surfaces of objects located in background scene. The 3D location of each point on this surface and its corresponding color, are computed using a Kinect sensor. Back Kinect $K_b$ delivers a measurement of the distances between its depth sensor and all objects in the background inside its field of view in real-time. These distances values are stored in a matrix of size 512x424 which is the pixels' size in depth image. Each pixel has distance value corresponding to the object point in the background that can be mapped from the pixel value to construct depthmap [7]. As each pixel of the depthmap represents a value for the corresponding point cloud, the calculated point clouds count 217088 points. Microsoft SDK (Software Development Kit) for Kinect v2 sensor, provides multiple functions for different data types such as mapping functions [8] that is used to map depth pixels to color pixels.

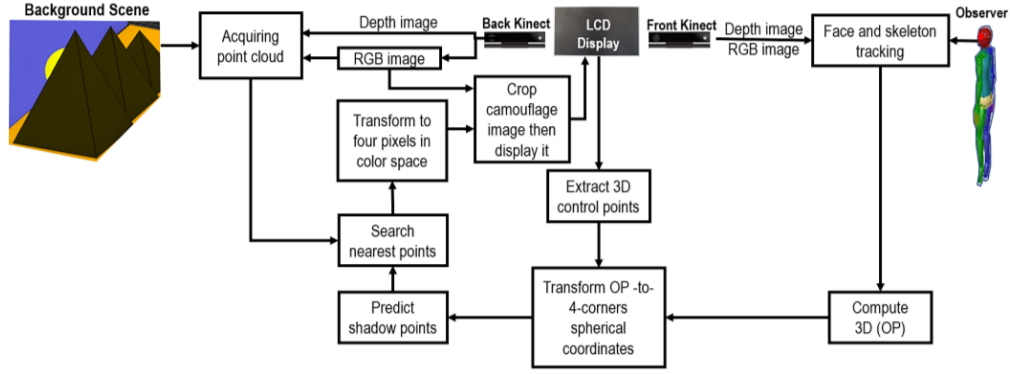

Fig. 1. Block diagram of the proposed camouflage system

The C++ implementation uses some code samples available in the Microsoft SDK. First, functions responsible for registration and alignment of RGB and depth cameras are called, hence pixels in color image goes with their corresponding depth pixels to overlap the depth and color images of the background. Second, synchronizing the processing time of different data streams, `MultiSourceFrames`, which represents a multi-source frame from the kinect sensor such as `IDepthFrame`, `IColorFrame` for depth and color data respectively. Third, converting between the three Kinect coordinate systems (3D camera space, 2D color space, and 2D depth space) using `ICoordinateMapper`, which represents the mapper that provides translation services from one-point type to another such as `MapDepthFrameToCameraSpace`, `MapDepthFrameToColorSpace` [8] for mapping pixel in depth coordinates to 3D corresponding point coordinates, and mapping pixel in depth coordinates to the corresponding pixel in color coordinates respectively. Fig. 2 shows the output point cloud of a background scene with defined and undefined (black) 217088 points. Practically, about 20% of the points in Fig. 2, are undefined or do not have corresponding color values in color image (i.e. the depth pixel actually has no projected pixel in the color image) and most of these points are located at the top and bottom of the depth image. Each point uses depth and color buffers that are filled with the coordinates ($x$, $y$, and $z$) of depth pixel instead of its value and its associated color value (red, green, and blue) respectively. If one of 217088 points does not have corresponding color value, then its color will be assigned to black.

### C. Predict Shadow points using Neural Networks

The control points of the LCD are defined to capture BG scene and track the OP at each of the four points. Then, coordinate transformations from 3D cartesian coordinates into spherical coordinate are performed as illustrated in Fig. 1. To explain the reason, let a predicted shadow point is $(x_p, y_p, z_p)$ at observation point $(x_o, y_o, z_o)$. In Fig. 3, the left image shows that there is only one shadow point for a set of multiple observation points located on the line connecting the corner and its shadow, i.e. the multiple OPs have the same azimuth and polar angles regardless of radial distance. Oppositely, the right image shows that at certain observation point, multiple shadow points can exist for only one corner depending on the structure of the background and the depth of object in BG that encloses this shadow point i.e. the multiple shadow points have the same azimuth and polar angles. And as the structure of BG frequently changes as the back kinect moves or in case of dynamic BG scene. Consequently, the 3D location of the corner shadow point changes under unpredicted rules. Then it is much easier to train ANNs using 3D shadow points in spherical coordinates than training them in 3D cartesian coordinates. An alternative way to use points in 3D cartesian coordinates in training NNs, is using the acquired point cloud of BG scene in estimating these unpredicted rules which is more complicated process and takes more time.

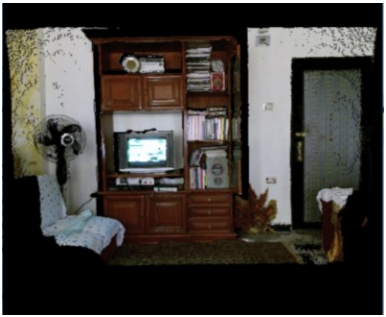

Fig. 2. Point cloud of a background scene captured by back kinect $K_b$

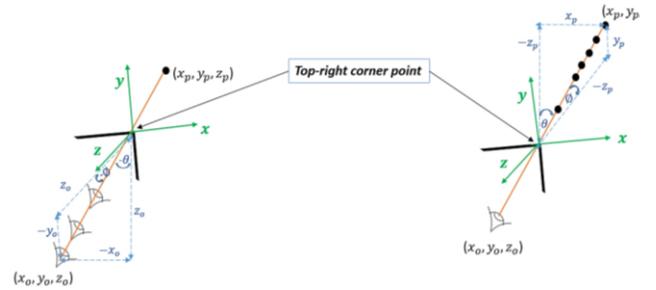

Fig. 3. Eye points and shadow points of top-right corner in Spherical coordinates and 3D Cartesian coordinates. Left image: multiple input tracked head point have one output depth BG point. Right image: one input tracked head point has multiple output depth BG point.

The camouflage problem is regarded as the problem of approximating a function from data or as function regression problem [9]. The goal is to produce a model that can predict shadow point of a corner ($\theta_t$, $\varphi_t$) for given observation point ($\theta_i$, $\varphi_i$). The input of the neural network is an azimuth angle $\theta_i$ or a polar angle $\varphi_i$, and the target is $\theta_t$ or $\varphi_t$ respectively.

*1) Experimental data*

Two experiments were performed on three backgrounds with different depths. Each experiment includes different camouflage data set type. In first experiment, the input and target are the angles of tracked 3D head points and depth points of corners' shadow respectively. The number of instances is 35. The data set has been obtained using the following procedures: the observer stands in front of the LCD display to (1) read the tracked parameters of the observer by the front kinect to compute the OP and then computing the input azimuth and polar angles, (2) assign the intersection point of each corner with the background scene, (3) mark these points in the RGB image of the background scene to get their pixels' locations, (4) search for their corresponding 3d locations in the constructed point cloud of BG scene, (5) convert from cartesian into spherical coordinates to get the azimuth and polar angles of each shadow point, and (6) repeat again for different observer's location. Fig. III, and 5 show the data set at two display corners which are at its top right and its bottom left respectively. The red points in the two figures represent azimuth angles $\theta$ (in degrees) input-target data, while the blue points are for polar angles $\varphi$ (in degrees) data set. The horizontal and vertical axis represent angles of tracked observation point and shadow depth points respectively.

The second experiment is carried out to overcome the error which results from the 1st experiment. Its input and the target are the output of 1st designed networks and correct measured angles of corners shadow. Fig. 6 illustrates the second experiment. A group of (16) true shadow points of TR and BL corners are shown in yellow color, while the false shadow points in red color. The arrow shows the direction from false to true point. We calculate the azimuth and polar angles of each true and false angles at TR and BL corners as shown in Fig. 7. It can be noted that the horizontal displacement of true angles, is the common error in this experiment, besides small vertical displacement error. This error results from the process of constructing point cloud, in which the points are laid out from depth image, starting at top left pixel, passing row by row from left to right. While, writing in the buffer should be from right to left instead of left to right (flipped horizontally).

We divide the 1st experiment input-target data set into 50%, 25% and 25% of the instances for training, validation and testing respectively. More specifically, 19 samples are used here for training, 8 for validation and 8 for testing. Random indices are generated for each of training, validation and testing instances.

*2) Multilayer perceptron*

Here a multilayer perceptron with a hyperbolic tangent hidden layer $L^{(1)}$ and a linear output layer $L^{(2)}$ is used. The multilayer perceptron must have one input (n=1), since there is one input variable; and one output neuron (m=1), since there is one target variable. One hidden layer $L^{(1)}$ will be enough. In order to draw the best network architectures that represent desired regression functions for solving the problem formulated above, different sizes ($s$) for the hidden layer are tested, and that providing the best generalization properties is adopted. In particular, the performance of three neural networks with 2, 3 and 4 hidden neurons are compared.

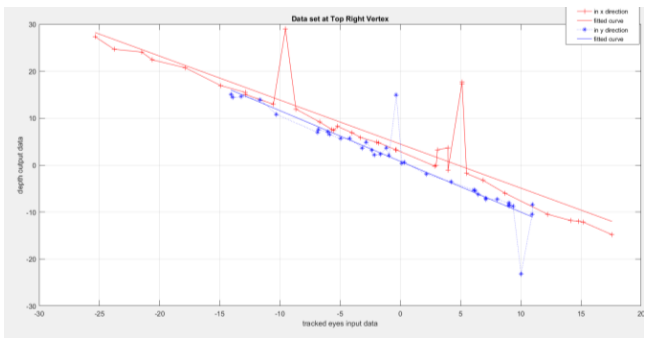

Fig. 4. Data set at top right vertex consists of 35 azimuth angles and polar angles in red and blue colors respectively.

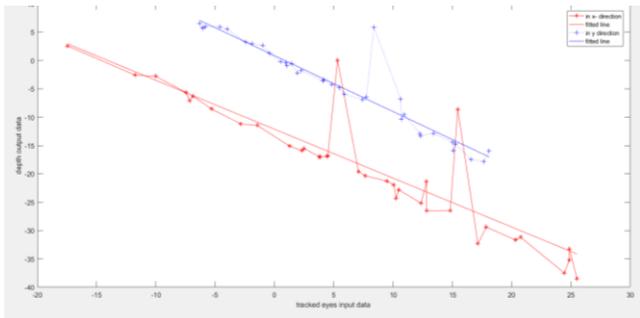

Fig. 5. Data set at bottom left vertex consists of 35 azimuth angles and polar angles in red and blue colors respectively.

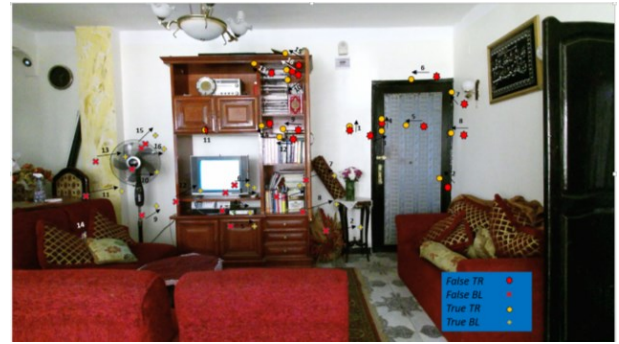

Fig. 6. One of the three backgrounds used in the 2nd experiment with colored points. Red and yellow points for false and true locations of corners, respectively.

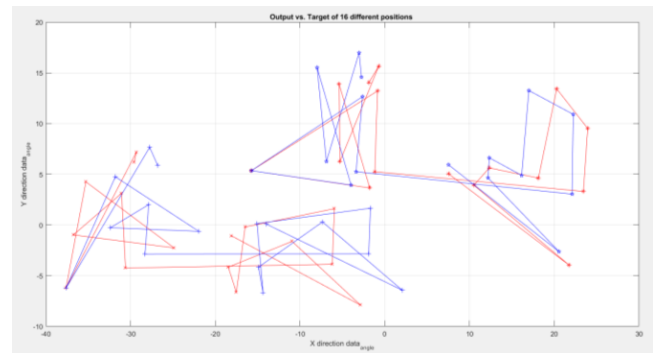

Fig. 7. Azimuth angles vary with polar angles of shadows of top right and bottom left vertices with both red false and blue true curves (locations of these points are shown in Fig. 6).

At each corner point, we design two neural networks, one for azimuth angle $\theta$ data set, and the other for polar angle $\varphi$ data set. TRX and TRY represent two neural networks at TR corner in $x$ direction ($\theta$ angles) and in $y$ direction ($\varphi$ angles) respectively. Other two networks at BL corner are BLX and BLY in $x$ and in $y$ direction respectively. Table I shows the training and validation errors, and regression parameters ($a$, and $b$) at different $s$ for each of required four networks. The regression parameters will be mentioned later in linear regression analysis. $E_T$ and $E_V$ represent the normalized squared errors made by the trained neural networks on the training and validation data sets, respectively. We can see in Table I, the training error decreases with the complexity of the neural network, but the validation error shows a minimum value for one multilayer perceptron in each network (yellow colored line). A possible explanation is that the lowest model complexity produces under-fitting, and the highest model complexity produces over-fitting that memorized the training examples, but it has not learned to generalize to new situations.

The graphical representation of the four network architectures (TRX, TRY, BLX and BLY) with optimal number of neurons in the hidden layer are depicted in Fig. 8. Each neural network has approximated the regression function well.

*3) Training algorithm*

The training algorithm is a quasi-Newton method with BFGS train direction and Brent optimal train rate. This method is one of first order algorithms which are recommended for solving function regression problems. The training subset is used for training the neural network. On the other hand, the error on the validation subset is monitored during the training process. And it normally decreases during the initial phase of training, as it does the training error. However, when the neural network begins to overfit the data, the error on the validation subset typically begins to rise. When the validation error increases, the training algorithm stops if it reaches 1000 epochs or when the improvement between two successive epochs is less than $10^{-12}$. Finally, the parameters at minimum validation error, are set to the neural network.

The presence of noise in the training data set which appears as (+) or (−) overshoots as shown in data sets in Fig. 3 and 4, makes the error function to have local minima. So, we repeat the learning process from several different starting positions during the training process to decrease the error function until the stopping criterion is satisfied. Table II shows the final values of some training results for each of the four designed neural networks in Fig. 9. Here $\|\zeta^*\|$ is the final parameter vector norm, $\|\nabla e(\zeta^*)\|$ the final gradient norm, $N$ the number of epochs and $T$ the CPU training time in a laptop Intel(R) Core(TM) i7-6500U. The evaluation $E_T$ history is shown in Fig. 9. Note that the plot has a logarithmic scale for the Y-axis.

*D. Linear regression analysis*

The regression analysis is performed between the network response and the corresponding targets for an independent testing subset. This analysis leads to 3 parameters for each output variable in each neural network. The first two parameters, a and b, correspond to the y-intercept and the slope of the best linear regression relating outputs and targets,

$$\theta_{out} = a + b\theta_{in} \quad (1)$$

where, $a = \dfrac{\sum_{q=1}^{Q} \theta_{out_q} \sum_{q=1}^{Q} \theta_{in_q}^2 - \sum_{q=1}^{Q} \theta_{in_q} \sum_{q=1}^{Q} \theta_{in_q}\theta_{out_q}}{Q \sum_{q=1}^{Q} \theta_{in_q}^2 - \left(\sum_{q=1}^{Q} \theta_{in_q}\right)^2}$

TABLE I. TRAINING AND VALIDATION ERRORS IN CAMOUFLAGE PROBLEM

| Four NNs | $s$ | $a$ | $b$ | $E_T$ | $E_V$ |
|---|---|---|---|---|---|
| TRX Neural N. | 2 | -0.00782583 | 1.03005 | 0.00797999 | 0.00902705 |
| | 3 | -0.0777348 | 1.00701 | 0.00553211 | 0.00223371 |
| | 4 | -0.0178394 | 0.999137 | 0.00011075 | 0.030471 |
| TRY Neural N. | 2 | -0.0125271 | 1.05538 | 0.00419984 | 0.00250758 |
| | 3 | 0.0707123 | 1.03706 | 0.00383026 | 0.00956217 |
| | 4 | -0.43111 | 1.03872 | 0.00023790 | 0.0208303 |
| BLX Neural N. | 2 | 0.0397477 | 0.984258 | 0.00399131 | 0.00791037 |
| | 3 | -0.0439661 | 0.981445 | 0.00374432 | 0.00519356 |
| | 4 | -4.74621 | 0.865134 | 0.00090126 | 0.0133052 |
| BLY Neural N. | 2 | -0.310875 | 1.13261 | 0.00412508 | 0.00620165 |
| | 3 | -0.0980712 | 1.08747 | 0.00239396 | 0.00585394 |
| | 4 | 0.18775 | 0.946407 | 0.00053426 | 0.03719 |

TABLE II. TRAINING RESULT VARIABLES

| Four NNs | $N$ | $T$ | $\|\zeta^*\|$ | $\|\nabla e(\zeta^*)\|$ |
|---|---|---|---|---|
| TRX Neural N. | 154 | 1 | 21.5587 | 1.33154e-08 |
| TRY Neural N. | 194 | <1 | 15.747 | 1.00812e-09 |
| BLX Neural N. | 575 | 2 | 54.3812 | 3.23758e-09 |
| BLY Neural N. | 687 | 1 | 8740.36 | 1.05947e-05 |

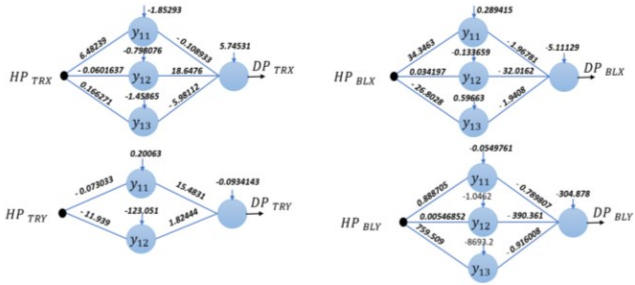

Fig. 8. Networks architectures with final neural parameters at TR and BL corners. Input is one of head point (HP) angles and output is one of depth point (DP) angles.

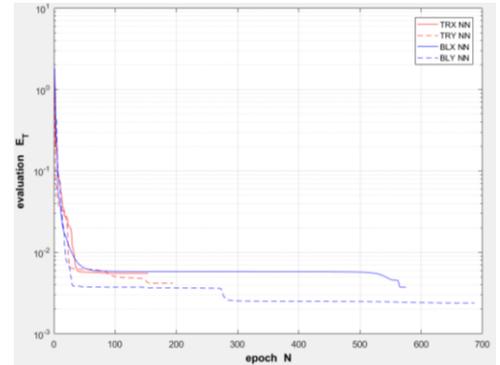

Fig. 9. Evaluation history for the camouflage problem.

$$b = \frac{Q \sum_{q=1}^{Q} \theta_{in_q} \theta_{out_q} - \sum_{q=1}^{Q} \theta_{in_q} \sum_{q=1}^{Q} \theta_{out_q}}{Q \sum_{q=1}^{Q} \theta_{in_q}^2 - \left(\sum_{q=1}^{Q} \theta_{in_q}\right)^2}$$

The third parameter, $R^2$, is the correlation coefficient between the outputs and the targets,

$$R^2 = \sum_{q=1}^{Q} \left(\theta_{out_q} - (a + b\theta_{in_q})\right)^2$$

$Q$ is number of instances. In case the input is polar angles, substitute $\theta$ with $\varphi$ in the previous equations.

If we had a perfect fit (outputs exactly equal to targets), the parameters values $a = 0$, $b = 1$ and $R^2 = 1$ means perfect correlation between the outputs from the neural network and the targets in the testing subset. Table I shows the linear regression parameters (a and b) for TRX, TRY, BLX and BLY networks. we can see that the neural network is predicting very well the entire range of Camouflage data. Indeed, a and b are very close to 0 and 1 respectively.

Fig. 10 shows the relation between the angles of input tracked eyes which are ($\theta_{in}$ and $\varphi_{in}$) and angles of output four shadow TR and BL corners which are $(\theta_{out}, \varphi_{out})_{TR}$ and $(\theta_{out}, \varphi_{out})_{BL}$ over the range $[-40°, 40°]$. The azimuth angle $\theta$ is defined around y-axis of kinect camera coordinate system (i.e. in the xz plane). While, polar angle $\varphi$ is defined around x-axis (i.e. in yz plane). Since kinect coordinates are transformed to corner points, the horizontal field of view (FOV) of depth sensor, 70°, which is smaller than FOV of RGB sensor (84.1°), defines azimuth range between $\pm 35°$. Also, polar angles range is equal to vertical FOV of RGB sensor, 53.8°, which is smaller than FOV of depth sensor (60°) as depicted with yellow dashed line. We can conclude that, the angles by which the observation point moves with respect to the origin of the front kinect located at corner point, is approximately equal in magnitude to the crossponding angles of the shadow point measured by back kinect located at that corner, but in opposite direction. This is practically exact at angles range $[-10°, 10°]$ as shown in Fig. 10. The curvature of the line out of this range, is due to the fact that kinect cameras do not rotate with the observer eyes or with the predicted shadow points, to center them in the middle of their images i.e. front camera has to rotate with him in order to see what exactly he see, simultaneously, back camera must rotate but in opposite direction. To minimize these errors which results from stationary kinect cameras, the line curvature has been created out of this angles range.

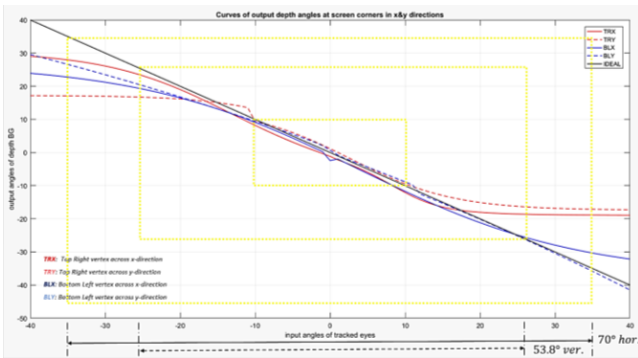

Fig. 10. Graphical representation of expressions for the camouflage model obtained by the four neural networks.

*E. Nearest neighbor search using KD Trees*

While observer moves in front of $K_f$ and looking towards the object-oriented display, the angles of the shadow points are computed by neural networks based on the angles of observation point ($\theta_{in}$, $\varphi_{in}$) in object coordinate systems at corner points. The computed angles $(\theta_{out}, \varphi_{out})_{TR}$ at top right corner, for example, at time instant t, is used to get the location of corresponding pixel in the RGB image of back kinect $K_b$. To get this location, it is required to search for the nearest angles $(\grave{\theta}_{out}, \grave{\varphi}_{out})_{TR}$ to computed $(\theta_{out}, \varphi_{out})_{TR}$ angles in the point cloud set of 217088 elements. Each element has information about the locations of depth pixel in depth space and its corresponding RGB pixel in color space; location of that point in the 3d coordinates systems and in spherical coordinates system of corner points, such as top right (TR) corner.

The technique used to search for nearest neighbor of $(\theta_{out}, \varphi_{out})$ is k-dimensional tree [11] which is a k-dimensional space used for partitioning and organizing point cloud elements. It is a special case of binary space partitioning (BSP) tree with a special rule that determines which splitting hyperplanes to use.

Fig. 11 shows a simple example of distributing and organizing of ten points in two-dimensional tree. These 2D points are (5.168°, −0.777°), (3.218°, −16.254°), (5.119°, −16.254°), (11.881°, −15.422°), (4.826°, −16.254°), (1.865°, −20.451°), (7.515°, −22.156°), (1.819°, −16.111°), (2.888°, −19.515°), and (5.166°, −17.116°) which represent the azimuth and polar angles at a corner. Then, it's desired to search for point with angles (11.881°, −15.422°) in point cloud of the ten points.

First, building this tree requires to choose a central point ($\theta_c$, $\varphi_c$) from the ten points and then divides the remaining points into two divisions depending on their $\theta$ values, such that the points which have $\theta$ value less than $\theta_c$ will be located in the left subtree, while points which have $\theta$ value greater than $\theta_c$ will be located in the right subtree. This central node is the root node which we'll say is at (4.826°, −16.254°); nodes one level down are (3.218°, −16.254°) at left where its $\theta$ component is less than 4.826°, and (5.166°, −17.116°) at right where its $\theta$ component is greater than 4.826°; nodes two levels deep are (5.168°, −0.777°) where its $\varphi$ is less than $\varphi$ component of its central node which is −16.254, (1.865°, −20.451°) where its $\varphi >$ −16.254, (5.119°, −16.254°) where its $\varphi < -17.116°$, and (7.515°, −22.156°) where its $\varphi > -17.116°$; nodes three levels deep are (1.819°, −16.111°) where its $\theta < \mathbf{1.865°}$, (2.888°, −19.515°) where its $\theta > \mathbf{1.865°}$, and (11.881°, −15.422°) where its $\theta > \mathbf{5.119°}$. Since the number of points in the cloud point are 217,088, the number of levels to build kd-tree is 18 levels ($2^{18} = 262,144$). After that is searching for point (11.881°, −15.422°) in the constructed tree. The search starts at the central point (4.826°, −16.254°) at its $\theta$ component if (11.881°>4.826°) then this point located in the right subtree. The new central node becomes (5.166°, −17.116°) so, if (−15.422°>−17.116°) then this point located in the right subtree. Then, the new central node becomes (5.119°, −16.254°) so, if (11.881°>5.119°) then this point located in the right subtree and all nodes in the left subtrees are ignored and the search algorithm ends. The number of visited nodes is 3 only instead of possibility of visiting 9 points. The counted range of visited nodes in a point cloud of 217,088 point to find nearest neighbor is 17 − 51 nodes.

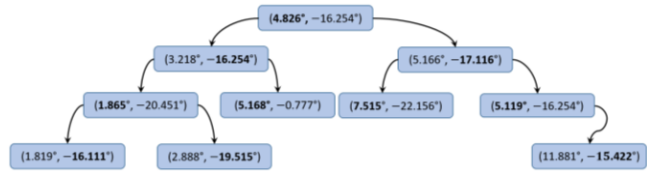

Fig. 11. The resulting k-d tree of these ten points.

Finally, Fig. 12 illustrates a practical show of the proposed camouflage system. The observer is tracked by the front Kinect sensor which is mounted on the LCD display and he holds a video camera near his eyes to record his frontal scene that includes the camouflaged image which is outputted on the LCD.

## IV. CONCLUSION AND FUTURE WORK

The proposed system has no constraints on location and direction of the hardware components (just Kinect constraints). The contributions of proposed Algorithm are (1) using point clouds of BG scenes to identify the occluded region in optical camouflage problem, also (2) predicting the shadow of the CO's corners on the background scene, (3) searching for the nearest neighbor in point clouds after transforming into spherical coordinates, and (4) tracking the observer's eyes to conceal the camouflaged object.

The proposed system overcomes some limitations of other systems. The limitations of retro-reflective projection technology (RPT) are the imperfect transparent of the covered object, the use of half-mirror in front of the observer's eyes, the existence of a projector in the observer side either on the ground or mounted on his head, and the projected image is not projected on the retroreflective material only but also on other surrounding objects. Drawbacks of projector-camera system, are the misalignment on the boundaries and severe distortion for 3D dynamic background. The limitation of the technique which hide a 3D object from many views simultaneously is that it must first capture the scene without the object to collect more than 10 images of the surroundings.

The main result of proposed system is that the OP azimuth and polar angles tracked by front Kinect are equal in magnitude to the azimuth and polar angles of the shadow point measured by back Kinect at that corner respectively, but opposite in direction. This is practically exact at angle range $[-10°, 10°]$. The accuracy decreases out of this range $\pm10°$. The results of this research suggest that the proposed system will be a promising practical solution for concealing objects.

In future work, optical camouflage needs more research contributions to solve more complicated problems such as: (1) camouflage 3D objects from different viewpoints using modern displaying technologies, (2) display of real-world 3D objects on hologram may help in solving camouflage problem in case of more than one observers see the camouflaged object. In that point, kinect can be used, as it has capabilities in constructing of 3D objects and also tracking up to six observers in real-time, (3) real-time camouflaging using movable front and back cameras, (4) sophisticated measurement of lighting conditions, surface reflections, and other BG properties in real time.

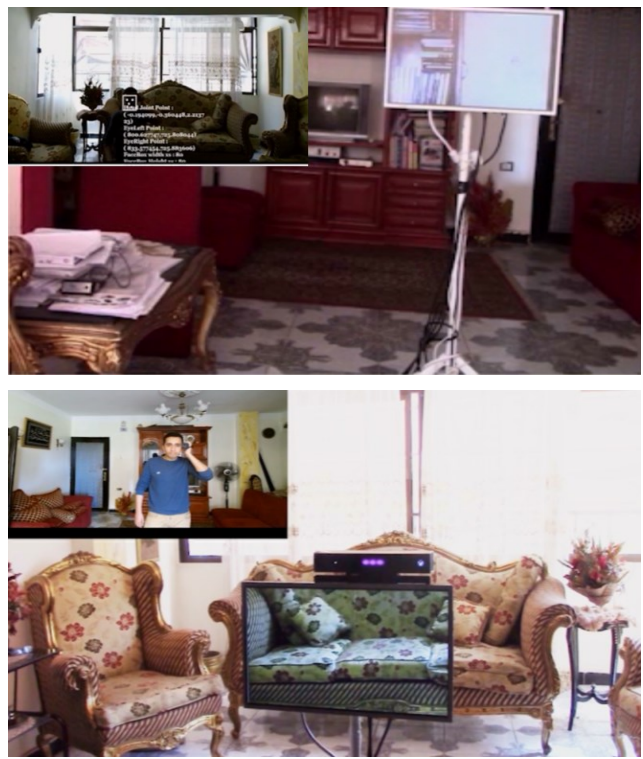

Fig. 12. Each image has two simultaneous images of tracked observer captured by front Kinect (top-left image) and image shows the camouflaged image displayed on LCD.